\documentclass[runningheads]{llncs}
\usepackage{graphicx}
\usepackage{amsmath, amssymb}
\usepackage{bm}
\usepackage{booktabs}

\newcommand{\acronym}{FaStR}

\newcommand{\titlename}{Factorized Structured Regression}

\begin{document}
\title{\titlename\\for Large-Scale Varying Coefficient Models}
%
%
\author{David R\"ugamer\inst{1,2} \and Andreas Bender\inst{1} \and Simon Wiegrebe\inst{1} \and Daniel Racek\inst{1} \and\\  Bernd Bischl\inst{1} \and Christian L. M\"uller\inst{1,3,4} \and Clemens Stachl\inst{5}}
\institute{Department of Statistics, LMU Munich \and Institute of Statistics, RWTH Aachen \and ICB, Helmholtz Zentrum Munich \and CCM, Flatiron Institute \and Institute of Behavioral Science and Technology, University of St.Gallen}
\maketitle              
\begin{abstract}
Recommender Systems (RS) pervade many aspects of our everyday digital life. Proposed to work at scale, state-of-the-art RS allow the modeling of thousands of interactions and facilitate highly individualized recommendations. Conceptually, many RS can be viewed as instances of statistical regression models that incorporate complex feature effects and potentially non-Gaussian outcomes. Such structured regression models, including time-aware varying coefficients models, are, however, limited in their applicability to categorical effects and inclusion of a large number of interactions. Here, we propose Factorized Structured Regression (FaStR) for scalable varying coefficient models. FaStR overcomes limitations of general regression models for large-scale data by combining structured additive regression and factorization approaches in a neural network-based model implementation. This fusion provides a scalable framework for the estimation of statistical models in previously infeasible data settings. Empirical results confirm that the estimation of varying coefficients of our approach is on par with state-of-the-art regression techniques, while scaling notably better and also being competitive with other time-aware RS in terms of prediction performance. We illustrate FaStR's performance and interpretability on a large-scale behavioral study with smartphone user data.

\end{abstract}

\section{Introduction}
\label{sec:intro}

From buying products online to selecting a movie to watch, 
recommender systems (RS) are part of our everyday life. RS are used to suggest those items that are most appealing to a given user based on the user's past preference data or the similarity of a user to other users. One big advantage of RS is their scalability, as they allow for modeling thousands of interactions, e.g., between users and items, and thereby facilitate individual recommendations in many dimensions (see, e.g., \cite{zhao2020recbole} for a recent implementation framework). Many RS can be represented as a regression model with the user and item as covariates. This makes it straightforward to include further features into the model and extend the method by other structural components. 

At the same time, the increasing amount of available data and the possibility to model increasingly complex data generating processes calls for efficient methods to fit flexible regression models on large-scale data sets with many observations and features. In the past, several advanced statistical regression models have been proposed to incorporate complex feature effects. One of the most common approaches are generalized additive models (GAMs), widely considered to be state-of-the-art (SotA) for statistical modeling \cite{Wood.2017.book}. These models allow the incorporation of time-varying feature effects and spatial effects, among others, and can also deal with non-Gaussian outcomes (see \cite{Wood.2017.book} for more details). While well-working adaptions of GAMs for large data scenarios exist \cite{Wood.2017}, both methodology and software reach their limits when modelling categorical effects or categorical interactions of several variables where features comprise hundreds or thousands of categories. 
An amalgamation of methods from RS and statistical regression can overcome the limitations of statistical regression models on large-scale data sets with many categorical effects and interactions. In this work, we combine smoothing approaches with factorization terms to overcome the limitations of varying coefficient models for categorical features with many factor levels. Our idea arises from the statistical analysis of a large-scale behavioral dataset analyzed in Section~\ref{sec:appl}. In this dataset, smartphone usage behavior of participants was tracked for several months. From this dataset, continuous user activity values aggregated over certain time periods can be obtained (e.g., average screen time in the morning, afternoon, evening and night) for every user and every activity. 
Domain experts are interested in various structured regression effects, such as the continuous activity levels over time. 
While standard regression software allows to fit some of these effects, several hundred activities and users make it infeasible to fit a model that learns interaction effects of users and activities or smoothly varying time effects for one or both of these categorical variables.  

\paragraph{Our contribution} We propose \titlename~(\acronym) for scalable varying coefficient models. This combined approach has the flexibility that has proven successful in additive regression models, while also being able to deal with high-dimensional categorical effects and interactions. More specifically, we 1) derive a general model formulation in \eqref{eq:proposal} to combine GAMs and factorization approaches, 2) derive a varying factorization interaction in Section~\ref{sec:facrep} that reduces the number of parameters and therefore computations by a factor of $(I+U)/(IU)$ for given numbers of category levels $I$ and $U$, and 3) propose an efficient implementation of this fusion approach that a) can reduce the storage cost by a factor of $1/(I+U)$ and b) circumvents computations quadratic in $I$ and $U$ by using stochastic optimization, an array reformulation, as well as dynamic feature encoding. In numerical experiments, we moreover show that our approach 4) leads to an estimation performance comparable with a SotA implementation and 5) has the desired computational complexity. Finally, we 6) demonstrate its interpretability and applicability to large-scale data sets, where it is also competitive with existing RS approaches. 

\section{Related Literature}

Multiple different RS have been proposed over the last years, many of them based on matrix factorization (MF) \cite{Srebro.2004,koren2009matrix} or collaborative filtering (see, e.g., \cite{thorat2015survey}). While recent methods increasingly rely on neural network-based factorization or recommendation (see, e.g., \cite{Wu.2021}), it remains debatable whether they yield superior results, e.g., with respect to performance and efficiency \cite{Jin.2021.understanding,rendle2020neural}. Factorization Machines (FM) represent another line of research which is closely related to MF. Initially proposed by \cite{Rendle.2010}, FM are based on a linear model problem formulation with pairwise (or order $d \geq 3$) interactions between all features. 
In particular, the formulation as regression model is the basis for extensions to other (prediction) tasks, with many different FM-type models having been developed in recent years (e.g., 
\cite{Blondel.2016,Hong.2019}).
An important influence (context variable) in RS is time. 
Various methods for controlling for short- and long-term temporal dynamics, cyclic patterns, drift or time decay exist \cite{Al.2017}. While short-term approaches either divide time into smaller periods or integrate time features into the factorization of the neighbourhood, long-term effects are accounted for by some form of distance calculations between the current and other designated time points. Some approaches also combine the factorization with the time context, and for instance assume smoothness in the factorization, e.g., for video pixel completion \cite{imaizumi2017tensor}. Specific time-aware methods include collaborative filtering with temporal dynamics \cite{koren2009collaborative}, dynamic MF \cite{Chua.2013}, temporal regularized MF \cite{Yu.2016}, or sequence-aware FM \cite{Chen.2020}. The common ground of these methods is to account for the time context in the factorization. 

\paragraph{Statistical approaches and interpretability} Several combinations of statistical approaches and RS have been proposed in past years. Already in 1999, \cite{condli1999bayesian} proposed a (Bayesian) generalized mixed-effects model as RS Likelihood approximation approach. Multiple probabilistic versions of matrix factorization have also been proposed (see, e.g., \cite{Shan.2010}). \cite{zhang2016} proposed the GLMix model that combines the idea of generalized linear models (GLMs; \cite{nelder1972generalized}) with RS for large-scale response prediction. 
Latest research puts strong emphasis on understanding and controlling of RS through improved model explainability \cite{Hada.2021}. 
Our work is most similar to RS approaches which facilitate interpretability by making connections to generalized additive models (GAMs) \cite{guo2020explainable,Zhuang.2021}. In contrast to our approach, however, past work does not include smoothing splines directly into the models, nor does it address varying coefficient models. An exception is the time-varying tensor decomposition by \cite{Zhang.2021.JMLR} which is inspired by varying coefficient models. While similar in motivation, their work does not focus on scaling aspects compared to classical approaches and only considers approximate varying coefficients with separately learned basis coefficients. Our approach implements the full varying coefficient model with exact single-varying coefficients as well as a doubly-varying coefficient with jointly-learned latent basis coefficients.

\section{Background}

We will first describe the necessary background on factorization approaches in RS, structured additive regression models, and introduce our notation.

\subsection{General Notation}

In the following, we use $Y \sim \mathcal{F}$ to denote a random outcome value (e.g., a rating) from distribution $\mathcal{F}$ and its observation $y \in \mathcal{Y} \subseteq \mathbb{R}$ for which the model generates a prediction $\hat{y}$. We reserve the indices $i \in \mathcal{I}$ and $u \in \mathcal{U}$ for two categorical features (exemplarily referred to as \emph{item} and \emph{user}) and $t \in \mathcal{T}$ for the context variable \emph{time} on a given time interval $\mathcal{T}$. The features associated with $i$ and $u$ are assumed to be binary indicator variables and are only implicitly referenced using their index. In Section~\ref{sec:proposal}, we however use an integer representation to introduce a memory-efficient storage representation. Other context features are summarized by $\bm{x} \in \mathbb{R}^p$. We use $b$, $w$ and $v$ to denote weights in the model that relate to items and users, and make their dependence explicit by indexing these weights correspondingly with $i$ and $u$. To distinguish between dependencies of categorical features and the (continuous) feature $t$, we highlight time-dependency by writing objects as functions of the time $t$.  We assume that we are given a dataset $\mathcal{D} = \{(y_{iu}(t), \bm{x}_{iu}(t))\}_{i \in \mathcal{I}, u \in \mathcal{U}, t \in \mathcal{T}}$ of total size $N$ and allow observations to be sparse, i.e., for $\mathcal{D}$ to be a true subset of $\mathcal{I}\times\mathcal{U}\times\mathcal{T}$. For matrix computations in later sections, let $\bm{A} \odot \bm{B} \in \mathbb{R}^{N \times a \cdot b}$ define the row-wise tensor product (RWTP) of matrices $\bm{A} \in \mathbb{R}^{N \times a}$ and $\bm{B} \in \mathbb{R}^{N \times b}$, i.e., a Kronecker product applied to every pair of rows of both matrices $[\bm{A}_{[1,]} \otimes \bm{B}_{[1,]} \ldots \bm{A}_{[N,]} \otimes \bm{A}_{[N,]}]$.  Further, for $a=b$, let $\bm{A} \bullet \bm{B} := (\bm{A} * \bm{B}) \bm{1}_b$, where $*$ is the Hadamard product and $\bm{1}_b \in \mathbb{R}^b$ a vector of ones. The operation defined by $\bullet$ can be exploited in models with Kronecker product structures such as array models for fast computation (see Section~\ref{sec:eff}).

\subsection{Model-based Recommender Systems}

The basic MF model generates its predictions as
\begin{equation} \label{eq:MF}
    \hat{y}_{iu} = \langle \bm{v}_{1,i}, \bm{v}_{2,u} \rangle
\end{equation}
using a dot product $\langle \bm{v}_{1,i}, \bm{v}_{2,u} \rangle = \bm{v}_{1,i}^\top \bm{v}_{2,u}$ of two latent factors $\bm{v}_{1,i}, \bm{v}_{2,u} \in \mathbb{R}^D$ from a $D$-dimensional joint latent factor space. After learning the mapping from each item $i$ and user $u$ to the respective latent factor vector, the dot product describes the interplay between user and item and is used to estimate the outcome $y$ (ratings). If the combination $\mathcal{I}\times\mathcal{U}$ is observed completely, common matrix decomposition approaches such as a singular value decomposition can be applied. If the matrix containing the ratings for all user-item combinations is sparse, missing values can be imputed. This, however, can be inaccurate and computationally expensive. The common alternative is to use \eqref{eq:MF} to only model $(i,u) \in \mathcal{K}\subset (\mathcal{I} \times \mathcal{U})$ in the set $\mathcal{K}$ of observed combinations. The solutions $\hat{y}$ can be found by least squares estimation where an additional $L_2$-penalty for $\bm{v}_i$ and $\bm{v}_u$ is typically added to the objective function \cite{koren2009matrix}. In order to account for systematic user- and item-level trends, biases are further added to \eqref{eq:MF}, yielding
\begin{equation} \label{eq:MFbias}
   \hat{y}_{iu} = \mu + b_i + b_u + \langle \bm{v}_{1,i}, \bm{v}_{2,u} \rangle,
\end{equation}
where $\mu$ is a global intercept representing the average rating, and $b_i, b_u$ represent the item and user tendencies. The latter two bias terms are again penalized using a ridge penalty. Together with this penalization, $b_i$ and $b_u$ can also be interpreted as a random effect for the item and user (see, e.g., \cite{zhang2016}).
 
\paragraph{Time-aware recommender systems} Contexts such as the location or time in which data has been observed can make a crucial difference (see, e.g., \cite{baltrunas2014experimental}). RS therefore often include a context dependency. One of the most common context-aware RS are time-aware RS \cite{campos2014time}. Time-aware model-based approaches assume the following relationship 
\begin{equation} \label{eq:timeFac}
      \hat{y}_{iu}(t) = \mu + b_i(t) + b_u(t) + \bm{v}_{2,u}^\top(t) \bm{v}_{1,i},
\end{equation}
where both biases and the dot product are time-dependent. The rationale behind a time-varying latent user effect is that users change their behavior over time, whereas influences of items should be time-independent \cite{koren2009matrix,campos2014time}. While time is often assumed to be continuous, categorical time-aware models are used if time information is represented as discrete contextual values. 

The time-varying latent user effect in \eqref{eq:timeFac} has a similar role as varying-coefficients in structured additive regression discussed in the following section.




\subsection{Structured Additive Regression and Varying Coefficient Model} \label{sec:proposal}

In statistical modeling, structured additive regression is a technique for estimating the relationships between a dependent variable (outcome value) and one or more independent variables (predictors, features). While the most common form of regression follows an additive structure as introduced in \eqref{eq:MFbias} and \eqref{eq:timeFac}, in particular including linear effects $\bm{x}^\top \bm{\beta}$ of features $\bm{x}$ or pairwise interactions, factorization terms are usually not present in classical regression models. Instead, to adapt models for complex relationships between features and outcome value, smooth, non-linear, additive terms $f(\cdot)$ of one or more features are included into the model. These terms are represented by a linear combination of $L$ appropriate basis functions. A univariate non-linear effect of feature $z$ is, e.g., approximated by $f(z) \approx \sum_{l=1}^{L} {B}_{l}(z) {w}_{l}$, where ${B}_{l}(z)$ is the $l$-th basis function (such as regression splines, polynomial bases or B-splines) evaluated at $z$ and $w_l$ is the corresponding basis coefficient. Similarly, tensor product basis representations allow for two- or moderate-dimensional non-linear interactions. It is also possible to represent discrete spatial information or cluster-specific effects in this way~(see, e.g., \cite{Wood.2017}). 

One important part of additive regression is the so-called \emph{varying coefficient} model \cite{Hastie.1993}. The rationale for these models is the same as for time-varying RS: effects of features in the model naturally vary over time. Therefore these models include effects $x f(t)$, such that the effect (coefficient) of $x$ is given by $f$ evaluated at time $t$, and $f$ is estimated from the data.
A special case is a varying coefficient $f_i(t), i=1,\ldots,I$ where a separate function $f_i$ is estimated for all $I$ levels of a categorical feature. Existing software to model varying coefficients with smooth time-effects is, however, not scalable to features with many categories. The bottleneck is an RWTP of the matrix of evaluated basis functions $\bm{B} := (B_1(t) \ldots B_L(t))$ and a (one hot-)encoded matrix for a categorical variable (e.g., item with $I$ levels). 

\paragraph{Computational complexity} Assuming equal number of basis functions $L$ for every smooth term $f_i$ in a varying coefficient model with $i=1,\ldots,I$ levels and $N$ observations, the storage required for the model matrix is $\mathcal{O}(N L I)$ and the computations $\mathcal{O}(N (L I)^2)$ (cf. \cite{Wood.2020}). Similar, for a model with an interaction effect of, e.g., item and user (with $U$ levels), the storage is $\mathcal{O}(N L I U)$ and $\mathcal{O}(N (L I U)^2)$.

\section{\titlename}

In order to address the computational limitations of statistical regression techniques, we will first introduce the general idea to obtain predictions from a \acronym~model and then go into more specific details and merits. We use the RS notation to define the model by means of a classical recommendation task with items, users, time context and an outcome $y$ such as a rating. As in a typical regression setting, further features $\bm{x}$ might exist that the modeler is interested in. We assume that the outcome $Y_{iu}(t)$ and all features $\bm{x}_{iu}(t)$ are observed on a grid for item $i = 1, \ldots, I$, user $u = 1,\ldots, U$, time $t = 1, \ldots, T$. While our approach also works for sparsely and irregularly observed data, we assume a grid of observations to simplify the notation. Conditional on the item, user, time and further features, $Y_{iu}(t)$ is assumed to follow a parametric distribution $\mathcal{F}$. We model the expectation of $Y_{iu}(t)$ as
\begin{equation}
\begin{split}
    \mathbb{E} (Y_{iu}(t)|\bm{x}_{iu}(t)) &= h(\eta_{iu}(t)),\\ 
    \eta_{iu}(t) &= \mu + b_i + b_u + b_{iu} + f^{[0]}(t) + f^{[1]}_{i}(t) + f^{[2]}_{u}(t) + f^{[3]}_{iu}(t)\\
    &\quad + \textstyle\sum_{o=1}^O g_{o}(\bm{x}_{iu}(t)).
\end{split}
    \label{eq:proposal}
\end{equation}
Here, $h$ is an activation or response function mapping the additive predictor $\eta_{iu}(t)$ onto the correct domain (e.g., $\mathbb{R}^+$ for a positive outcome variable). All terms indicated with $b$ are (regularized) bias terms. $\mu$ is a global bias term, $b_i$ an item-specific bias, $b_u$ a user-specific bias and $b_{iu}$ an item-user-specific one. Terms denoted by $f$ are smooth non-linear functions of time $t$ represented by (penalized) basis functions. These include a global trend $f^{[0]}$, a subject and activity trend, $f^{[1]}$ and $f^{[2]}$, respectively, and a joint trend $f^{[3]}$. Additional covariates $\bm{x}$ can be modeled using other (smooth) functions $g_o$. In Supplemental Material~\ref{app:penopt} we provide further details on smoothness penalties and optimization of the model.

The model in \eqref{eq:proposal} can be seen as an alternative notation for a varying coefficient model, or also as a time-aware RS with additional exogenous terms. As $\mathcal{F}$ is not required to be Gaussian, it has, however, a more general applicability (e.g., binary, count or interval data). What further distinguishes \eqref{eq:proposal} from existing approaches is the smoothness assumption of terms denoted with $f$, combined with the efficient implementation of terms $f^{[1]}$, $f^{[2]}$ and a factorization assumption for $f^{[3]}$. These aspects are explained in more detail in the following. 

\subsection{Varying Factorized Interactions} \label{sec:facrep}

For high-dimensional data, such as the mobile phone data in our example, estimating the 2- or 3-way interaction terms is computationally not feasible. We thus propose to define $\eta_{iu}(t)$ in \eqref{eq:proposal} using latent factorization representations. We therefore decompose the discrete interaction term(s) into an inner product 
\begin{equation}
    b_{iu} = \langle\bm{v}_{1,i},\bm{v}_{2,u}\rangle = \textstyle\sum_{d=1}^D v_{1,i,d} \cdot v_{2,u,d}
\end{equation}
with $D \ll \min(I,U)$ latent factors, resulting in the estimation of $D\cdot (I + U)$ instead of $I\cdot U$ parameters. If $I=U=1000$ and $D=5$, for example, this reduces the number of parameters by a factor of $100$ from $10^6$ to $10^4$. While this is the common approach to model interactions in factorization approaches, we here propose to proceed in a similar fashion to model time-dependent interactions and approximate time-varying interactions by a factorization approach:
\begin{equation} \label{eq:vfac}
    f^{[3]}_{iu}(t) \approx \tilde{f}^{[3]}(t,\bm{V}_{1,i},\bm{V}_{2,u}) = \textstyle\sum_{l=1}^L B_l(t) \textstyle\sum_{d=1}^D v_{1,i,l,d} \cdot v_{2,u,l,d},
\end{equation}
where $\bm{V}_{\cdot,\cdot} \in \mathbb{R}^{L \times D}$ are matrices with rows corresponding to the $L$ basis functions for one categorical effect  and columns to the $D$ latent factors. In other words, we approximate the interaction of the smooth effect $f(t)$ of $t$ and categorical variables $i,u$ by a product of the non-linear basis $\bm{B}$ of dimension $L$ and the two latent matrices $\bm{B}^\top (\bm{V}_{1,i} \bullet \bm{V}_{2,u})$, which can be computed efficiently for all $N$ rows. The representation via latent factors requires the estimation of $L\cdot D\cdot (I + U)$ instead of $L\cdot D \cdot I\cdot U$ parameters (a multiplicative reduction of $(I+U)/(IU)$ in parameters and computations). This principle is general and can be applied to various types of additive effects, also of two or higher dimensions such as tensor-product splines or Markov random field smooths (see \cite{Wood.2017.book}). 

\paragraph{Penalization} In order to enforce smoothness of the varying coefficients in the time-dimension, a quadratic Kronecker sum penalty $J$ can be added to the loss function \cite{Wood.2017.book}. In a similar manner, we can promote smoothness of the latent factors $\bm{V}_{iu} = (\bm{V}_{1,i} \bullet \bm{V}_{2,u}) \in \mathbb{R}^L$ in our adaption using a symmetric difference penalty matrix $\bm{P}_V \in \mathbb{R}^{L \times L}$. $\bm{P}_V$ penalizes the time-dimension of the factorized varying-coefficients, where the penalized differences (its entries) depend on the chosen basis $B$. We further allow for an $L_2$-regularization of the latent factors in the $i$- and $u$-dimension, yielding
\begin{equation}
    J = \lambda_{t} \cdot \bm{V}_{iu}^\top \bm{P}_V \bm{V}_{iu} + \lambda_{iu} \cdot (||\bm{V}_{1,i}||_F + ||\bm{V}_{2,u}||_F),
\end{equation}
where $\lambda_{t}$ controls the smoothness of the non-linearity in the direction of the time $t$, $||\cdot||_F$ is the Frobenius norm, and $\lambda_{iu}$ the regularization for items and users.

\subsection{Efficient Implementation} \label{sec:eff}

While the previous section allows to efficiently model (smooth) interactions of two or more categorical features with many categories, the factorization is not a solution for coefficients $f^{[1]}$ and $f^{[2]}$ varying with a single category. Many use cases also require to estimate one effect for each of the levels of a categorical (interaction) effect. One bottleneck if $I$ and/or $U$ is large, is the computation of their dummy-encoded design matrices. For example, for item the matrix $\bm{X}_I$ of size $N \times I$ contains binary entries indicating which observation (rows) belong to which item category (column), and analogous for a user matrix $\bm{X}_U$. 
Second, the execution of operations involving such large matrices is another computational bottleneck. Computations become even more challenging if the model includes interactions, resulting in $\mathcal{O}(NIU)$ storage and $\mathcal{O}(N(IU)^2)$ computations (cf.~Section~\ref{sec:proposal}). These interactions are created by calculating the RWTP between both matrices, i.e., $\bm{X}_I \odot \bm{X}_U \in \mathbb{R}^{N \times I\cdot U}$. To circumvent creating, storing and processing $\bm{X}_I$ and $\bm{X}_U$ as a whole, we propose two simple yet effective implementation tricks explained in the following.\\

\noindent \textbf{Stochastic optimization} The first bottleneck in computations of varying coefficient models at scale is the number of observations $N$. We therefore implement \acronym~in a neural network and thereby can make use of stochastic gradient descent optimization routines with mini-batches of size $M \ll N$. This reduces the original cost of computations from $\mathcal{O}(N(IU)^2)$ to $\mathcal{O}(EMIU)$ where $E$ is the number of model updates. It also allows us to leverage high-performance computing platforms such as TensorFlow \cite{TensorFlow} that support GPU computations. \\

\noindent \textbf{Array computations} The second bottleneck is computing the RWTP of two- or higher-dimensional interaction terms. Our proposal is to use an array reformulation that does not require to compute the RWTP design matrix in the first place. More specifically, a two-dimensional interaction effect $(\bm{X}_I \odot \bm{X}_U) \bm{w}$ with weights $\bm{w}$ can be equally represented by 
\begin{equation} \label{eq:array}
(\bm{X}_I \bm{W}) \bullet \bm{X}_U,    
\end{equation}
where $\bm{W} \in \mathbb{R}^{I \times U}$ is a matrix of weights with the $i$th row and $u$th column being the weight for the interaction of the $i$th level in $\bm{X}_I$ and the $u$th level in $\bm{X}_U$. By using \eqref{eq:array} instead of a plain linear effect, we circumvent the construction of the large RWTP and the storage cost is reduced from $\mathcal{O}(N IU)$ to $\mathcal{O}(N(I+U))$ without increasing the time complexity (as the $\bullet$ operation for $M$ observations is neglectable with $\mathcal{O}(MU)$ compared to the matrix multiplication with $\mathcal{O}(MIU)$). This array formulation can also be defined for higher-order interactions \cite{Currie.2006}.\\

\noindent \textbf{Dynamic feature encoding} Although array computations can reduce the storage problem notably by not constructing the RWTP in the first place, a third bottleneck is storing the large dummy-encoded matrices $\bm{X}_I$ and $\bm{X}_U$ themselves. We circumvent this extra space complexity, by evaluating categorical features dynamically during network training and only constructing the one-hot encoding for categorical features on a given mini-batch $m$. Thereby, only a matrix $\bm{X}^{(m)}_I$ of size $M \times I$ needs to be loaded into memory and the full $N \times I$ matrix is never created explicitly. This effectively reduces the storage from $\mathcal{O}(NIU)$ to $\mathcal{O}(N)$ (exactly $2N$ for two integer vectors). While this potentially results in redundant computations as it will create the encoding for a specific observation multiple times (if the number of epochs is greater than 1), deterministic integer encoding is cheap. Hence, the resulting computation overhead is usually neglectable and both forward- and backward pass can make use of the sparse representation.

When evaluating the two matrix operators (matrix product and $\bullet$) in \eqref{eq:array} sequentially, encodings can again be created dynamically and the largest matrix involved only contains $\max(I,U)$ instead of $I\cdot U$ columns. 


\section{Numerical Experiments}

Our numerical experiments investigate 1) whether \acronym~can estimate (factorized) varying coefficients as proposed in Section~\ref{sec:facrep} with performance comparable to other SotA methods, and 2) whether the model presented in \eqref{eq:proposal} can be initialized and fitted with constant memory scaling w.r.t. the number of factor levels. 
In addition, in the Supplementary Material~\ref{app:further} we investigate  whether \acronym~can recover GAMs in general using our implementation techniques. Details on the data generating processes for each simulation study, method implementation and the computing environment are given in the Supplementary Material~\ref{app:sim}. In the first experiment, we compare the estimated and true model coefficients using the mean squared error (MSE) and the estimated vs. the true functions $f$ using the squared error, integrated and averaged over the domain of the predictor variables (MISE). We repeat the data generating process 10 times to account for variability in the data and model estimation. 

\subsection{Estimation of Factorized Smooth Interactions} \label{sec:simfac}

We first investigate how well our models can recover smooth factorized terms from Section~\ref{sec:facrep} of two categorical variables $j$ and $u$, with 4 and 5 levels, respectively, i.e., 20 different smooth effects. We use a relatively small number of levels to be able to fit all possible interactions also in a classic structured regression approach. We define one of the two true factorized smooth effects as a varying coefficient term, i.e., $f_{1,iu}(t)$, and one stemming from an actual factorization, i.e., $f_2(t, \bm{V}_{1,i}, \bm{V}_{2,u})$. We use the Bernoulli and Gaussian distributions and investigate factorized terms in the distributions' mean for different data sizes $N\in\{500, 1000, 2000\}$. While the true model is generated using three latent factor dimensions, we additionally investigate a model with six dimensions to see how the misspecification of the latent dimension size influences estimation performance. FaStR is trained using a batch size of $250$ with early stopping on 10\% of the training data and a patience of $50$ epochs.  

\begin{figure}[ht]
    \centering
    \includegraphics[width=\columnwidth]{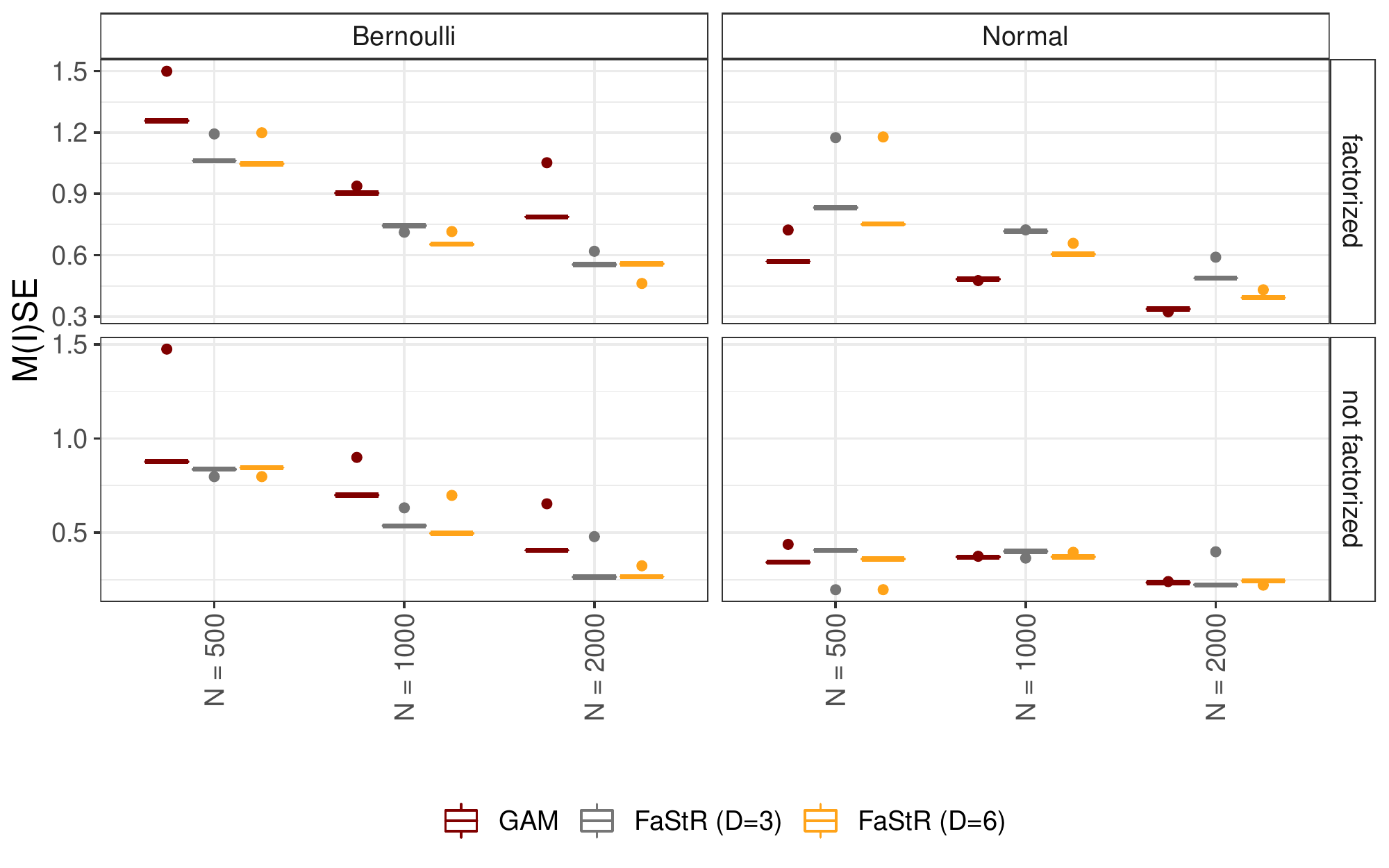}
    \caption{Comparison of M(I)SE values for the estimated partial effects of different methods (colors) for factorized and non-factorized smooth terms (x-axis), different distributions (columns) and different data sizes (rows). Values $> 1.5$ are set to the value $1.5$ to improve readability. 
    }
    \label{fig:estimation_factorized}
\end{figure}

\paragraph{Results} All results show that FaStR can estimate varying coefficients equally well 
compared to a classic GAM estimation routine.
Figure~\ref{fig:estimation_factorized} shows the resulting M(I)SE values for all data settings. Note that the GAM implicitly assumes as many latent dimensions as there are factor levels, but can also shrink single smooth functions to zero. In cases where it is feasible to fit smooth effects for every factor level (as is the case here), GAM can thus be seen as gold standard. For the case of a normal distribution, we observe that GAM yields better results, but also that the performance of FaStR converges to the one from GAM with increasing number of observations. For the Bernoulli case, our approach benefits from the optimization in a neural network and even outperforms the classic GAM estimation routine which requires a multiple of observations compared to a Gaussian setting for good estimation results (as, e.g., also found in \cite{Ruegamer.2020}). 
\subsection{Memory Consumption}

Finally, we compare the memory consumption of our implementation against the SotA implementation for big additive models (BAM) in the R \cite{R} package \texttt{mgcv} \cite{mgcv} for an increasing number of category levels (20, 40, 60, 80) when using a categorical effect or a varying coefficient based on the representation proposed in Section~\ref{sec:eff}, $N \in \{1000, 2000, 4000\}$ and optimization as in Section~\ref{sec:simfac}. While the improvement in memory consumption is expected to be even larger when using factorized terms instead of interaction terms with weights for each category combination, we do not use factorization in this experiment as there is no equivalent available in software for additive regression. Additionally, we also track the time when running FaStR for $10$ epochs to see if there are notable changes in the time consumption for varying data generating settings.

\begin{figure}[ht]
    \centering
    \includegraphics[width=0.95\columnwidth]{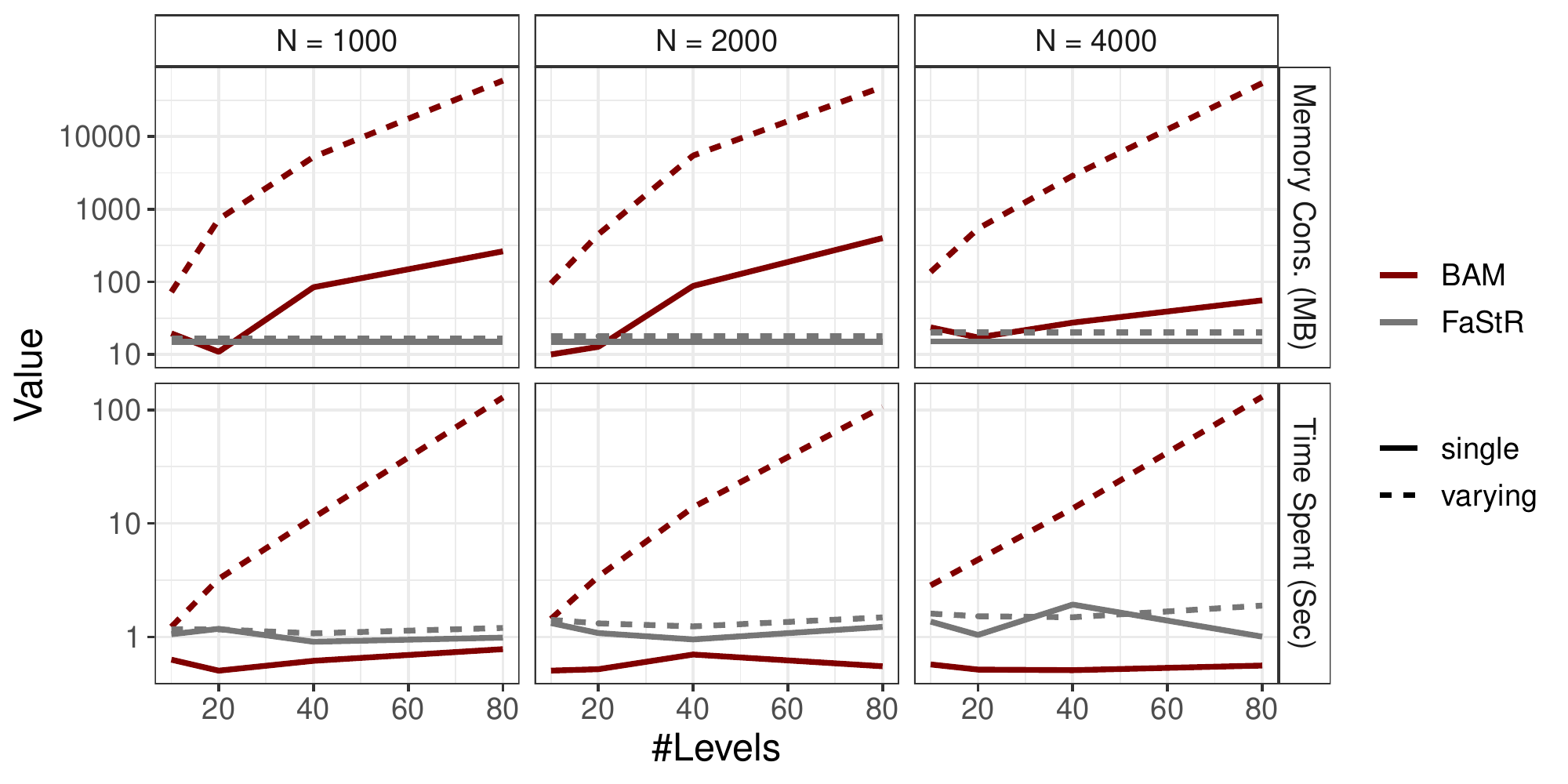}
    \caption{Memory and time consumption (y-axis; $\log_{10}$-scale) comparison between the SotA big additive model (BAM) implementation and our method (in different colours) for an increasing number of categorical levels (x-axis) of a factor effect (single) and varying coefficient term (varying) for different data sizes (columns).}
    \label{fig:memory}
\end{figure}

\paragraph{Results} Figure~\ref{fig:memory} visualizes the results for all different settings and compares run times and memory consumption of the two methods for factor variables (single) and varying coefficient effects (varying). Results show that FaStR has both, almost constant time and memory consumption while the SotA method requires exponentially more memory for growing numbers of factor levels (as the whole encoded matrix must be loaded into memory). These results confirm that our implementation works as intended to allow for the estimation of varying coefficient models in large-scale settings.

\section{Benchmarks} \label{sec:bench}

Although the focus of this work is to provide scalable and interpretable regression models, prediction performance of our models is also of (secondary) interest. We aim for a similar performance compared to SotA time-aware RS techniques, yet without the ambition to outperform these methods. We use the \emph{MovieLens 10M movie ratings} benchmark data set \cite{Harper.2015}, which is sparse in terms of user-item combinations, with \emph{items} corresponding to rated movies. In addition, we benchmark models on a subset of the densely observed \emph{PhoneStudy behavior} data set \cite{Stachl2020}, analyzed in more detail in Section~\ref{sec:appl}. In both cases, we use single train-test splits (90\%/10\% and 70\%/30\%, respectively) and evaluate the models predictive performance with the root mean squared error (RMSE). The different characteristics of both data sets are given in Table~\ref{tab:desc}. 
\begin{table}[ht]
\centering
  \caption{Descriptive statistics of the two benchmark data sets.}
  \label{tab:desc}
  \begin{tabular}{lcc}
     & Movies  & Phone Study  \\
    \midrule
  \# Observations ($N$) & $\approx 9$ m & $\approx 8.7$ m  \\
    \# Users ($U$) & 69,878 & 342 \\
    \# Movies/Activities ($I$) & 10,677 & 176 \\
    \# Unique Time Points ($T$) & $\approx$ 6.5 m & 348 \\
  \bottomrule
\end{tabular}
\end{table}
\vspace*{-0.6cm}
\paragraph{Methods} As comparison we use Bayesian timeSVD and Bayesian timeSVD++ flipped, two variations of the SVD++ method \cite{koren2008factorization}, a latent factor model whose key innovative feature is the incorporation of implicit user information. Both Bayesian timeSVD and Bayesian timeSVD++ flipped have been extended to be time-aware \cite{koren2009collaborative} and optimized by Gibbs sampling using a Bayesian reformulation \cite{rendle_difficulty_2019}. 
Bayesian timeSVD++ flipped integrates both implicit user and item information and has been reported to be the best-performing model among multiple SotA methods in a recent benchmark study \cite{rendle_difficulty_2019}.
The second variation, Bayesian timeSVD, is still a time-aware latent factor model, yet it does not incorporate implicit user or item information. As we are mainly interested in the performance of the proposed time-varying coefficient model, the Bayesian timeSVD provides a much fairer comparison with FaStR as it does not include the aforementioned types of implicit information. We use tuning parameter settings as given in \cite{rendle_difficulty_2019} for the two benchmark methods (i.e., we use the already tuned models).
\begin{table}[h]
\centering
\begin{small}
  \caption{Benchmark results based on the RMSE on the respective test data set for both benchmark data sets (columns) for different methods (rows). Best results are in bold with the best results of the respective other method underlined.}
  \label{tab:freq}
  \begin{tabular}{lll}
      &Movies & PhoneStudy \\
    \midrule
   timeSVD & 0.872  & 0.089 \\
    timeSVD++ flipped \quad \quad & \textbf{0.856} & \underline{0.087} \\
    \acronym~($D=10$) & 0.984  & \textbf{0.076}  \\
    \acronym~($D=3$) & 0.975  & 0.080  \\
    \acronym~($D=1$) & \underline{0.890}  & 0.087  \\
    \acronym~(w/o $f^{[3]}$) & 1.027  & 0.093  \\
  \bottomrule
\end{tabular}
  \end{small}
\end{table}
We compare these two models against our method as proposed in \eqref{eq:proposal} and thereby not only test its predictive performance, but also its capability to scale well to high-dimensional data sets. We do not tune FaStR extensively, but perform a small ablation study by testing different latent dimensions ($D\in 1,3,10$) for the factorized varying coefficient term $f^{[3]}$ and by excluding the whole term. All models use early stopping based on a validation data split with the same proportion as the train-test split.

\paragraph{Results} Table~\ref{tab:freq} shows the performance of all methods. Interestingly, FaStR is competitive with the SotA timeSVD approaches, even though the premise of this paper was merely to develop a scalable variant of the varying coefficient model, not to propose a method with SotA performance on RS tasks.
\section{User Behavior Phone Study} \label{sec:appl}
\begin{figure}[!ht]
    \centering
\includegraphics[width=0.85\columnwidth]{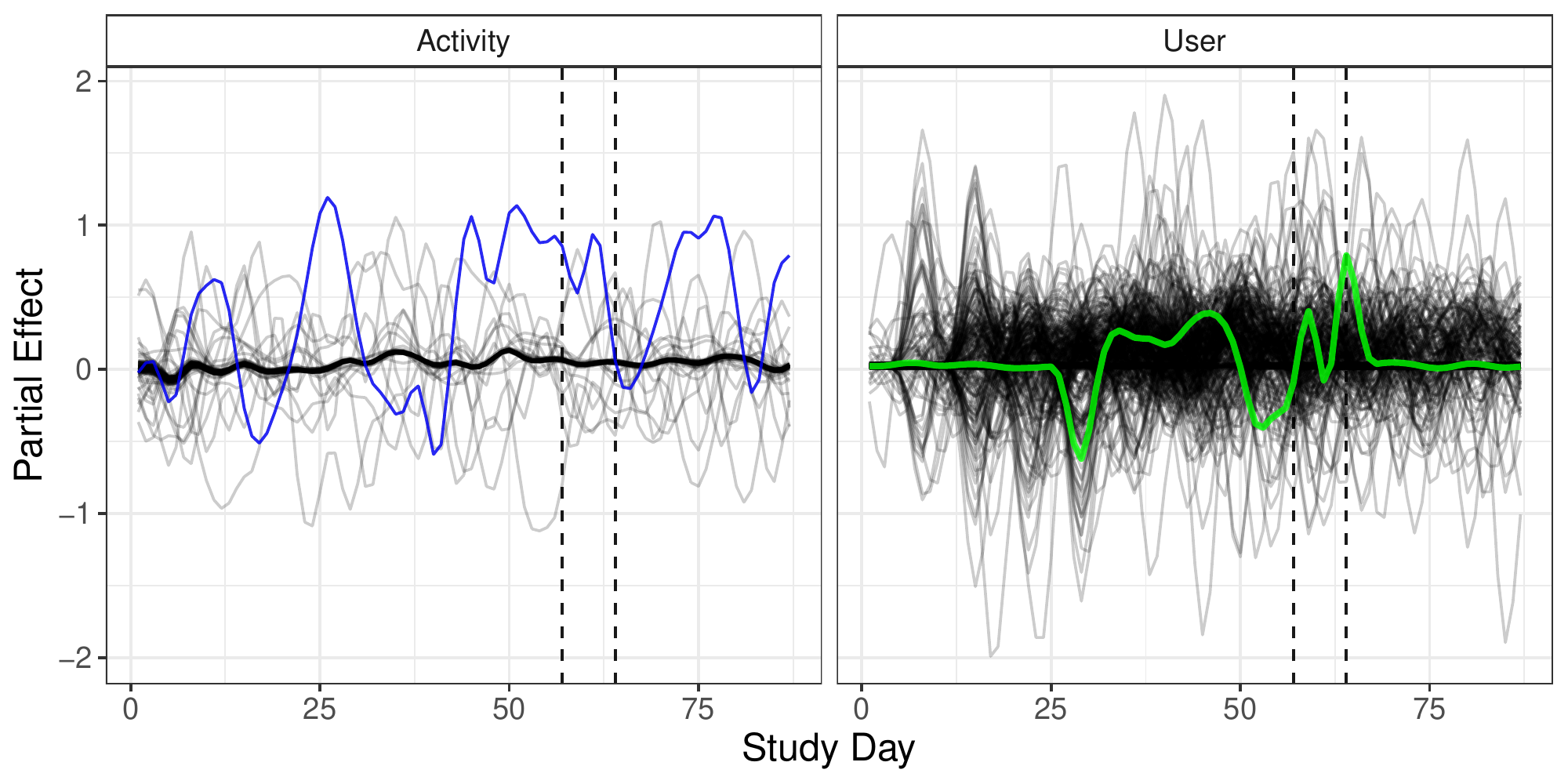} 
\caption{One dimension of the latent time-varying coefficients for different activities (left plot with ``locked screen'' in blue) and users (right plot with exemplary user in green) over 87 study days (x-axis). Vertical lines show Christmas and New Year's Eve.}
    \label{fig:varcoef}
\end{figure}
We finally turn to the motivating case study. A more detailed description of the study and data set can be found in the Supplementary Material~\ref{app:phone}. To analyze the activity levels (a value between $0$ and $1$, indicating the intensity of the activity in the given aggregation window) of participants in the study, we model the expected activity levels using user and activity effects, their interaction based on a factorization, an effect of the day of the week (Mon - Sun), the time of the day (6-hour windows), a factorized interaction of users and weekday as well as users and daytime, and a smooth time-dependent study day effect varying by user and/or activity. All factorizations use a three-dimensional latent space.

\paragraph{Results}

Results are generally plausible and in line with prior expectations. While various model effects are examined in the Supplementary Material, we here briefly analyze the factorized varying coefficient interaction by analyzing its latent factors. One of the dimensions is depicted in Figure~\ref{fig:varcoef}. Most activities follow a global pattern (darker concentration of lines), while a few show very specific sequences. For example, the ``locked screen'' event is observed less in the days around New Year's Eve. At the user level, no general pattern is visible (in part due to the different starting dates of the participants), but we still observe sudden changes in activity around holidays such as Christmas.  

\section{Conclusion and Outlook}

In this work we presented an amalgamation of structured statistical regression and RS to allow for large-scale varying coefficient models with many categorical levels. For this, we leveraged factorization approaches combined with an efficient neural network-based implementation. Empirical results confirm the efficacy of our proposal. 
In order to make the proposed approach as flexible as commonly used statistical regression software, we used a model-based point of view and cast the approach as a generalized additive model.  An interesting future avenue and additional advantage of formulating the RS model as structured additive regression is the possibility to extend this approach to distributional regression, in order to account, e.g., for aleatoric uncertainty in the data generating process. 

\section*{Acknowledgement}
This work has been partially supported by the German Federal Ministry of Education and Research (BMBF) under Grant No. 01IS18036A.




\bibliographystyle{splncs04}
\bibliography{literature}

\appendix

\section{Penalization and Optimization} \label{app:penopt}

The proposed model can be optimized by (restricted) Maximum Likelihood with negative log-likelihood as loss function and additional difference penalty terms $J_0, \ldots, J_3$ for model terms $f^{[0]},\ldots,f^{[3]}$ that penalize differences in neighboring basis coefficients to enforce smoothness (see, e.g., \cite{Wood.2017.book} for more details). In order to determine the amount of smoothness (the amount of penalization), every penalty term is controlled with a respective smoothing parameter $\lambda_0,\ldots,\lambda_3$. Next to tuning these parameters, a possible option is to opt for equal amount of smoothness of all involved non-linear functions. This can be realized by setting the \emph{degrees-of-freedom} $\text{df}_j$ for each smooth term $j\in \{0,\ldots,3\}$ to the same value $\text{df}$. As all involved terms are penalized linear smoothers \cite{Buja.1989}, we can exploit the one-to-one mapping between $\text{df}_j$ and $\lambda_j$, and efficiently calculate the different $\lambda_j$ values corresponding to $\text{df}_j \equiv \text{df}$ using the Demmler-Reinsch Orthogonalization (see Appendix B.1.1 of \cite{ruppert2003semiparametric}). Given the different $\lambda$ values, varying coefficient models such as \eqref{eq:proposal} are convex optimization problems that can be solved efficiently (see, e.g., \cite{chu2013distributed}).

\section{Further Numerical Experiments} \label{app:further}

\subsection{General Estimation Performance} \label{sec:simest}

The goal of the experiments in this subsection is to show that our framework is able to represent effects of GAMs and works well when applying implementation techniques from Section~\ref{sec:eff} together with neural-based optimization routines. To this end, we simulate GAM models for different data sizes $N\in\{1000, 2000, 4000\}$ from different distributions (Bernoulli, Beta, Gaussian, Poisson) with five non-linear smooth terms, a tensor product smooth, a varying coefficient effect with a categorical variable $i$ with 10 levels, and an interaction term of two categorical variables $j$ and $u$, each with 10 levels. \acronym~is trained using a batch size of $50$ with early stopping on 10\% of the training data and a patience of $100$ epochs. 

\begin{figure}[ht]
    \centering
    \includegraphics[width=0.85\columnwidth]{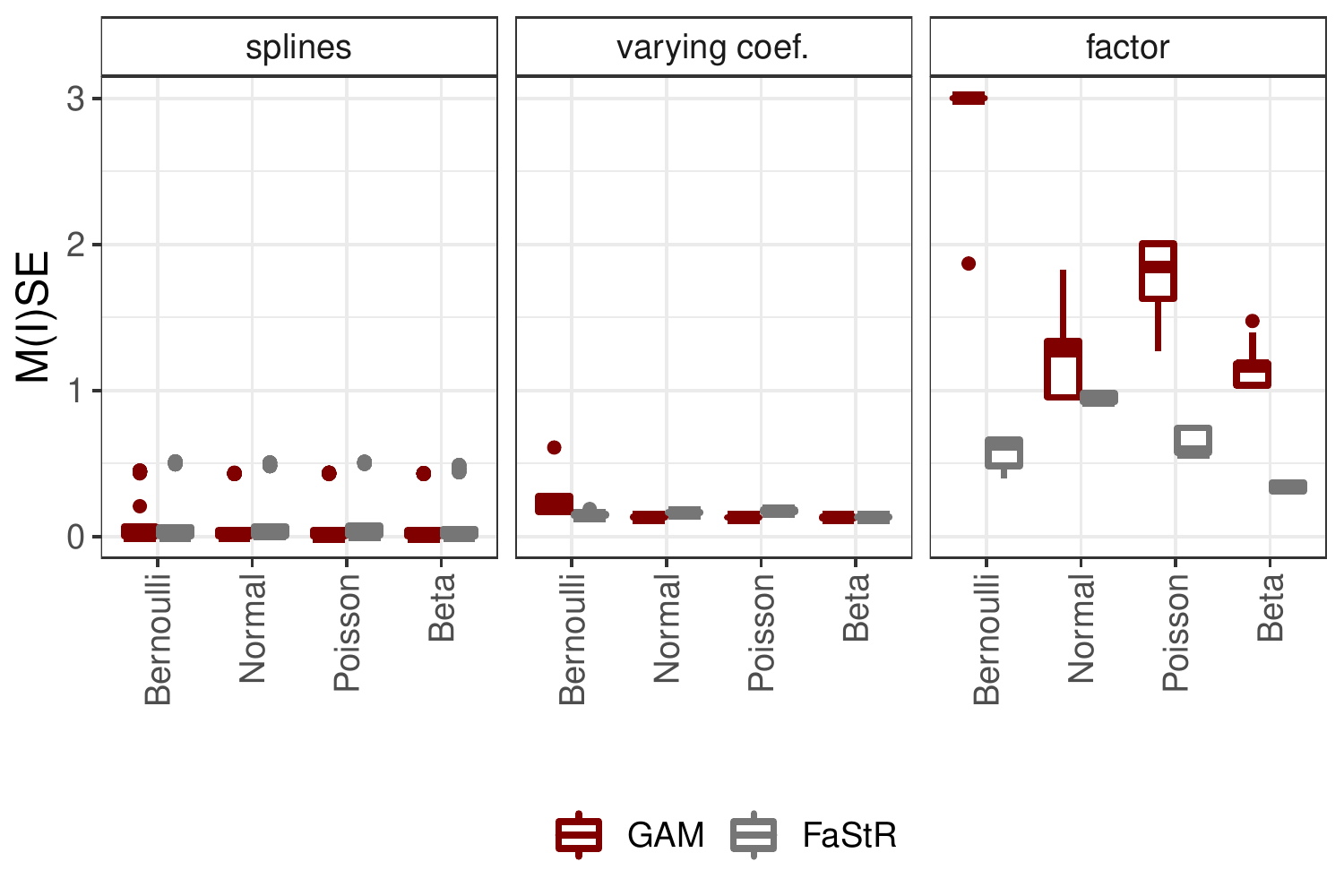}
    \caption{Comparison of model fits of the SotA GAM implementation (in red) and our method (in gray) for different model terms (columns), different distributions (x-axis) and corresponding M(I)SE values visualized by boxplots over 10 runs and different numbers of observations. Outliers (values larger 3) are set to 3.}
    \label{fig:estimation_nonlinear}
\end{figure}

\paragraph{Results} All results indicate that \acronym~can estimate the different effects and distributions equally well when compared to a SotA estimation technique for GAMs \cite{mgcv}). Figure~\ref{fig:estimation_nonlinear} shows the resulting mean (integrated) squared error (M(I)SE) values\footnote{See Supplementary Material~\ref{app:sim} for a detailed definition.} for all simulation settings, indicating that FaStR's estimation of the splines, factor effects and varying coefficients works well and even seems less vulnerable to difficult data settings (here the estimation of categorical variable effects for the Bernoulli distribution).

\section{Details on Numerical Experiments} \label{app:sim}

We here describe the details how data is generated in our numerical experiments. For details on the computing environment see Section~\ref{app:details2}.\\


In all simulations, we use the same data generating process to create the additive predictor $\eta_{iu}(t)$. Depending on the distribution $\mathcal{F}$ to be simulated, we transform $\eta_{iu}(t)$ with the respective $h$-function to generate the distribution's mean $\theta_{iu}(t) = h(\eta_{iu}(t))$ and draw outcome values according to $Y_{iu}(t) \sim \mathcal{F}(\theta_{iu}(t))$. The additive predictor $\eta$ itself is generated by first drawing numerical features from a standard uniform distribution and categorical features using a balanced design with $I$ (or $U$) different categories and $\lfloor I/N \rfloor$ ($\lfloor U/N \rfloor$) observations per category. Based on the numeric values, univariate splines are generated using the functions $f_{uni,1}(x) =  -(x/4)^5$, $f_{uni,2}(x) = \log(x^2)/10$, $f_{uni,3}(x) = \sin(3x)$, $f_{uni,4}(x) = (\sin(3x)\cdot 0.1 + 1) \cdot I(x<0) + (-2x + 1)/4 \cdot I(x \geq 0)$ with indicator function $I(\cdot)$ and $f_{uni,5}(x) =  -x \tanh(3x) \cdot \sin(4x) / 4$. Bivariate tensor-product splines are generated using the function $f_{bi}(x,y) = (\pi^{0.3} \cdot 0.4) \cdot (1.2 \cdot \exp(-(x - 0.2)^2/0.09 - (y - 0.3)^2/0.16) + 0.8 \cdot \exp(-(x - 0.7)^2/0.09 - (y - 0.8)^2/0.16))$ and the smooth part of varying coefficient splines using one of the following functions (depending on the category of the factor variable): $f_{vc,1}(x) = \cos(3x)$, $f_{vc,2}(x) = \tanh(3x)$, $f_{vc,3}(x) = -(x/4)^3$, $f_{vc,4}(x) = \cos(3x-2) \cdot (-x/3)$, $f_{vc,5}(x) = \exp(x/2) - 1$, $f_{vc,6}(x) = (x/2)^2$, $f_{vc,7}(x) = \sin(x)\cos(x)$, $f_{vc,8}(x) = \sqrt{|x|}$, $f_{vc,9}(x) = -(x/4)^5$, $f_{vc,10}(x) = \log(x^2)/100$.

We evaluate all non-linear functions by taking the point-wise MSE and integrate over the whole function domain. More specifically, given a function $f(t)$ and estimate $\hat{f}(t)$ of feature $t\in\mathcal{T}$ with sampled values $t_1,\ldots,t_n$, the MISE is calculated by $\mathbb{E} \int_{\mathcal{T}} (\hat{f}(t) - f(t))^2 dt \approx n^{-1} \sum_{i=1}^n \Delta_i (\hat{f}(t_i) - f(t_i))^2$, where $\Delta_i$ are integration weights. We do this on the $n$ simulated training data points as the goal is to evaluate estimation performance.

\section{Details Phone Study} \label{app:phone}

\subsection{Motivation}

In the behavioral sciences and psychology in particular, the understanding and prediction of behavior is one of the main goals. For a long time, it has been assumed that behavior can be seen as a function of a person's internal factors (e.g., personality traits, past experiences) and external factors (e.g., perceived environment) \cite{Lewin1943}. However, as trivial as this assumption might seem, it has proven very difficult for researchers to investigate this assumption empirically \cite{Baumeister2007}. The currently ongoing digital revolution in behavioral sciences is driven by the increasing availability of large quantities of digital behavioral data which can be collected directly with consumer devices. High-dimensional, fine-grained mobile sensing data has emerged as one of the most promising sources of information to study human behavior and experience, unobtrusively, in the wild, and at very large scale \cite{Harari2016}. This development requires new methods to handle large-scale data while also yielding interpretable models. Here, we demonstrate the usefulness of \acronym~in a first analysis of human behavior inferred from a high-dimensional mobile sensing data.

\subsection{Data set and model specification}

The PhoneStudy dataset \cite{Stachl2020} consists of behavioral metrics that were obtained with smartphone sensing. The dataset features various activities (mobility, calendar events, music listening, calling, power charging, app use and more). Here we use a subset of the dataset that includes up to 87 days of sensing and that was collected during a study period from October 28th, 2017 to January 22nd, 2018. After pre-processing, we obtain activities from 342 participants in Germany in 6-hour windows ([0-6], [6-12], [12-18], [18-24]) throughout each day of the study period.


\subsection{Results}

Based on the model's estimated effects, WiFi events (connected/disconnected/disabled), airplane mode, Bluetooth (off) and power charging are the top categories with highest overall (i.e., user-independent) activities, while certain apps show the lowest overall activity.
\begin{figure}
    \centering
    \includegraphics[width=0.8\columnwidth]{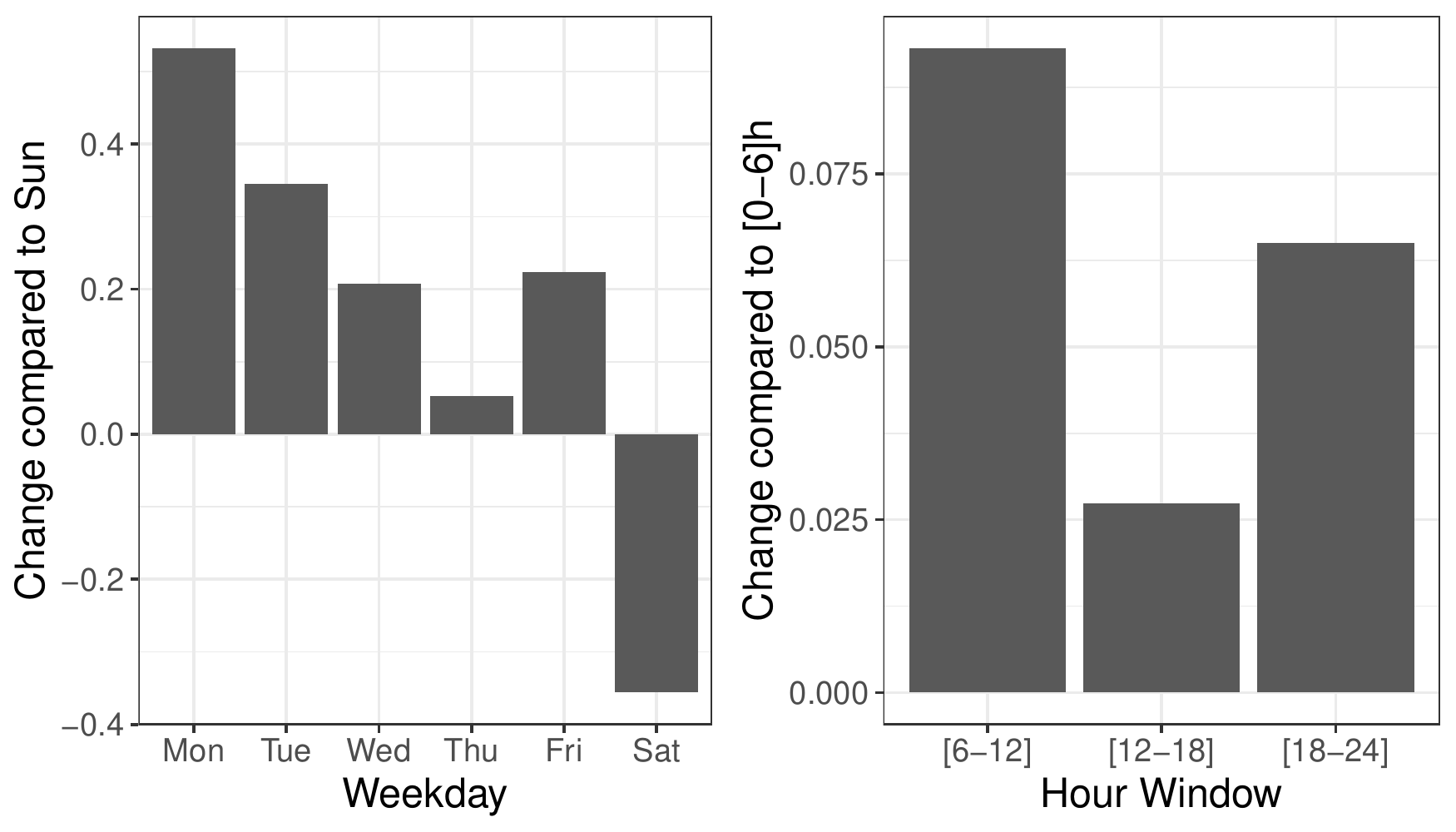}
    \caption{Partial effects of the day of the week (compared to Sunday; upper row) and the time of the day (compared to hours 18-24) on the mean (left column) and standard deviation (right column).}
    \label{fig:weekhour}
\end{figure}
Results further show that activity of mobile phone usage notably decreases towards the end of the week (Figure~\ref{fig:weekhour}) with activity on Saturday being the lowest, and users are least active in the [0-6]h window (i.e., during sleep), closely followed by the window [12-18]h (e.g., during work/school). Moreover, the results of our factorization approach show that some activities are specific to certain time points (e.g., phone charging is done more often at the beginning or end of the day, while airplane mode is more often active during the night). All of these effects are plausible and demonstrate the usefulness of \acronym~to model contextualized human behavior.

\section{Computing Environment} \label{app:details2}

For comparability reasons, \acronym~is implemented in R and only runs on a CPU (as it is the case for the reference software for GAMs \cite{mgcv}). The details on computing environments used for both numerical experiments and benchmark studies are as follows.
For all numerical experiments we used the R version 4.1.2, Python 3.7, TensorFlow 2.5.0rc0 on a personal computer with Intel(R) Core(TM) i7-8665U CPU @ 1.90GHz, 15.3 GB main memory and Ubuntu 20.04.3 LTS 64-bit as operating system. Timings and memory consumptions were tracked using \texttt{bench} \cite{Hester.2020}.


While we run our approach on the Phone Study data on the same environment as described in Section~\ref{app:details2}, we use a cloud server with Intel(R) Xeon(R) Gold 6148 CPU @ 2.40GHz, 92.8 GB main memory and Ubuntu 18.04.3 LTS for the Movies data. We also use the latter environment to run the timeSVD++ models on both data sets.

\section{Outlook: Distributional Regression}

The presented model in \eqref{eq:proposal} can be generalized as follows:
\begin{equation}
    Y_{iu}(t)|\bm{x}_{iu}(t) \sim \mathcal{F}({\theta}_{1,ij}(t), \ldots, {\theta}_{K,ij}(t)), \label{eq:dist}
\end{equation}
where the $K$ parameters $\theta_k$ of the distribution $\mathcal{F}$ can be described by a structured additive (factorized) predictor
\begin{equation} 
    \theta_{k,ij}(t) = h_k(\eta_{k,ij}(t)), k=1,\ldots,K. \label{eq:safm}
\end{equation}
The model described in \eqref{eq:dist} and \eqref{eq:safm} is a distributional regression model which relates some or all distribution parameters to available features in $\eta_k, k=1,\ldots,K$ \cite{Ruegamer.2020,Kneib.2021}. An interesting property to investigate in future research is how DR can account for aleatoric uncertainty. This can, e.g., be useful to check whether the model's uncertainty varies over different context variables time. 

\end{document}